\documentclass{article}
\usepackage{spconf,amsmath,epsfig}
\usepackage{url}
\usepackage{graphicx}
\usepackage{subfig}
\usepackage{amsfonts}
\usepackage{amssymb}
\usepackage{stmaryrd}
\usepackage{cite}

\title{$K$-Nearest Neighbor algorithm coupled with Logistic Regression in medical Case-Based Reasoning systems.}
\subtitle{Application to prediction of access to the renal transplant waiting list in Brittany.}

\author{Boris Campillo-Gimenez\sthanks{Corresponding author: B. Campillo-Gimenez, Inserm U936 Facult\'e de M\'edecine, Rue du Pr L\'eon Bernard 35043 Rennes cedex - Tel: +33(0)299284215 - E-mail address: boris.campillo-gimenez@univ-rennes1.fr}, Wassim Jouini, Sahar Bayat, Marc Cuggia}
\address{Unit\'e Inserm U936, IFR 140, Facult\'e de m\'edecine, Universit\'e Rennes 1, \\ 2 avenue du Professeur L\'eon Bernard 35043 Rennes Cedex 9, France.}

\begin{document}
\maketitle

\begin{abstract}

\par \emph{Introduction.} Case Based Reasoning (CBR) is an emerging decision making paradigm in medical research where new cases are solved relying on previously solved similar cases. Usually, a database of solved cases is provided, and every case is described through a set of attributes (inputs) and a label (output). Extracting useful information from this database can help the CBR system providing more reliable results on the yet to be solved cases. 
\par \emph{Objective.} For that purpose we suggest a general framework where a CBR system, viz. K-Nearest Neighbor (K-NN) algorithm, is combined with various information obtained from a Logistic Regression (LR) model. 
\par \emph{Methods.} LR is applied, on the case database, to assign weights to the attributes as well as the solved cases. Thus, five possible decision making systems based on K-NN and/or LR were identified: a standalone K-NN, a standalone LR and three soft K-NN algorithms that rely on the weights based on the results of the LR. The evaluation of the described approaches is performed in the field of renal transplant access waiting list. 
\par \emph{Results and conclusion.} The results show that our suggested approach, where the K-NN algorithm relies on both weighted attributes and cases, can efficiently deal with non relevant attributes, whereas the four other approaches suffer from this kind of noisy setups.  The robustness of this approach suggests interesting perspectives for medical problem solving tools using CBR methodology.

\end{abstract}


\section{Introduction}
\label{sec:Intro}
\newcommand{\E}{\mathbb{E}}
\renewcommand{\P}{\mathbb{P}}
\newcommand{\1}{{\rm 1\hspace*{-0.4ex}
\rule{0.1ex}{1.52ex}\hspace*{0.2ex}}}
\newtheorem{lemma}{Lemma}
\newtheorem{theorem}{Theorem}
\newtheorem{assumption}{Assumption}
\newtheorem{fact}{Fact}  
\newtheorem{definition}{Definition}

\subsection{Case Based Reasoning for Medical Applications}
\label{sec:CBR}
\par Case-based reasoning (CBR) is a problem-solving paradigm emerging in medical decision-making systems \cite{bichindaritz_advances_2011}. Instead of relying solely on general knowledge of a problem domain, CBR utilizes the specific knowledge of previously experienced, concrete problem situations - also referred to as \textit{cases} - to tackle new ones \cite{pantazi_case-based_2004}.

\par More specifically, CBR methodology defines a general \textit{CBR cycle} composed of four steps centered around a case database \cite{aamodt_case-based_1994}. First, the decision making process needs to identify, among the solved cases, those that seem to be the most similar to the considered unsolved case.
Then, solve the new case relying on the knowledge extracted from the most similar solved cases.
The third step consists in evaluating the suggested solution for the new case. Finally, if the solution is found satisfactory, the decision making process usually stores the part of the experiment likely to be useful for future problem solving. 

\par CBR in biology and medicine has found one of its most fruitful application areas and appears particularly suited to designing decision making tools in the field of Health sciences \cite{bichindaritz_case-based_2006-1}. Indeed,  Medicine appears as a highly intensive-data field where it is advantageous to develop systems capable of reasoning from pre-existing cases such as from electronic health record repositories for instance.


\subsection{Problem Definition and Objectives}
\label{sec:PbObj}
\par This paper focuses on the two first steps of the CBR cycle, viz. retrieve and reuse solutions from previously experienced situations. 
\par Knowledge in CBR systems consists of cases. Each case is a problem description linked to its solution. For solving new problems, the decision making process requires to select relevant cases, by measuring similarity of common characteristics between the new and the previously experienced cases \cite{bergmann_experience_2002}. 

\par In accordance with the traditional CBR view, the knowledge database contains cases, which consist in a problem-specific definition and construction. Thus, there are as many case bases as problems to be solved. Bergmann \textit{et al.} overcome that problem by introducing concept of utility \cite{bergmann_utility-oriented_2001}. Similarity measures are not directly computed from the problem descriptions of new and previously experienced cases, they are computed with the description of their utility; utility description being specifically defined in accordance with the solution needed.  

\par Statistical analyses and regression modeling could be useful to introcuce utility description in CBR systems, by converting medical data sources \emph{- or data bases -} into medical case bases. Regression models contain a part of knowledge which may be used to define utility description of cases and to perform problem-specific measures of similarity. The paper precisely consists of such an illustration by the formal definition and evaluation of a traditional CBR retrieval algorithm \lq{}\textit{the $K$-Nearest Neighbor ($K$-NN) algorithm\rq{}} coupled with a logistic regression model.
\vspace{0.1cm}
\par The rest of the paper is organized as follows : First, Section~ \ref{sec:scope} specifies the scope the paper. Then, Sections~\ref{sec:model} and \ref{sec:LPBLR}  respectively detail the decision making model and the considered learning process. Section~\ref{sec:EPR} focuses on the implementation, evaluation and interpretation of the suggested methodology. Finally, Section \ref{sec:Related} discusses related works and perspectives.

\section{Scope of the Study}
\label{sec:scope}
\subsection{Domain Application and Data Source}
\label{subsec:DomApp}

\par To carry out this work, we used data from the French Renal Epidemiology and Information Network (REIN) registry \cite{couchoud_renal_2006} related to renal replacement therapies (RRT) for end-stage renal disease (ESRD), and data from the \textit{Agence de la Biom\'edecine}, the French national agency of organ transplantation for registration on the waiting list of kidney transplantation. 
\par Registration on the waiting list is a medical decision based on medical factors in accordance with French medical guidelines that do not really need automated decision-making support. Nevertheless, those data and their domain application were chosen for several reasons: 
\begin{itemize}
\item Data come from a national registry that confirms the data quality by the French \textit{Comit\'e National des registres} agreement. 
\item Many studies showed that the selection criteria on the waiting list diverge from one center to another, and that access to the renal transplant waiting list is influenced by both medical and non medical factors \cite{bayat_medical_2006}.
\item Recent studies showed that it is possible to predict access to the waiting list relying on some of these factors \cite{bayat_modelling_2008,bayat_comparison_2009}.
\item Our main objective is a methodological essay on combination of CBR retrieval algorithm with logistic regression, and not the implementation of a medical decision support.
\end{itemize}

\subsection{Study Population and Data Collection}
\label{subsec:SPop}
\par The study population consists of every incident ESRD patients in Brittany, limited to those who started an RRT (peritoneal dialysis or hemodialysis) between January the 1st, 2004 and December the 31th, 2008. Patients who received a preemptive transplant and patients who came back on the waiting list after a first transplant have been excluded.
\par Registration status on the transplant waiting list was computed relying on the date of the first RTT as well as the date of registration on the waiting list. Only patients recorded on the waiting list within 12 months after inclusion on the REIN registry have been considered as registered patients.  
\par  A set of description factors have been defined according to data availability of the REIN database and the renal transplant scientific literature  \cite{fritsche_practice_2000,oniscu_equity_2003,jeffrey_comparison_2005,bayat_medical_2006,ravanan_variation_2010}. All factors have been dichotomized, i.e., reduced to a binary value. Three categories of factors likely to be related to registration on the transplant waiting list have been studied:
\begin{itemize} 
\item Social and demographic factors: sex, age and current occupation at the first RRT. 
\item Clinical and biological factors at the first RRT: existence of hypertension, diabetes, chronic respiratory failure, chronic heart failure, ischemic heart disease, heart conduction disorder or arrhythmia, positive serology (HCV, HBV, HIV), liver cirrhosis, disability, past history of malignancy and hemoglobin as $<$11 g/dl and  $\geq$11 g/dl. 
\item Factors related to medical care: ownership of nephrology facility where the first RRT were performed (private or public), follow-up in institution performing transplantation, type of first RRT (hemodialysis or peritoneal dialysis), urgent versus planned first dialysis session and first catheterization.
\end{itemize}
\par Due to missing data ($\geq$10\%), some factors potentially related to registration on the waiting list have not been considered either for statistical analyses or CBR algorithms: distance from patient’s residence to the transplantation department, smoking status, body mass index, vascular comorbidities and serum albumin level.

\section{Decision Making Model}
\label{sec:model}

\subsection{Decision Making Process and mathematical notations}
\label{subsec:DMP}
\par We depict, in this subsection, the overall mechanism designed to predict patient accessibility to renal transplant waiting list. Upper case notations refer to vector (or a set of vectors, viz., a matrix) whereas lower case notations refer to scalar real variables\footnote{An exception is made for the scalar parameter $K$ of the $K$-NN algorithm for the sake of consistency with the literature.}. Curved notations denote sets of elements.
\par For the sake of generality,  Let $\pi$ refer to the decision making process considered hereafter.  Moreover, let $\mathcal{C}_L$ refer to a set of labeled cases, viz. patients,  and let $\mathcal{C}_U$ refer to a set of new analyzed cases. We aim at designing a decision making process that maps new cases to previously solved (i.e., labeled) cases.
\par We consider two possible classes: as matter of fact, a patient is either registered in the renal transplant waiting list or not. Consequently, the labels are assumed to be binary.  let $y_{p}\in\{0,1\}$ denote the label assigned to patient $p\in \mathcal{P_L}$, where $\mathcal{P_L}$ refers to the set of patients considered in $\mathcal{C}_L$.
\par The set of \textit{cases} consists, in either case-sets -labeled, $\mathcal{C}_L$, or not $\mathcal{C}_U$- of a set of patients, $\mathcal{P}_L$ (or  $\mathcal{P}_U$ respectively), and two sub-sets: $\mathcal{A}$ and ${V}_L$ (or ${V}_U$ in the case of $\mathcal{C}_U$) named respectively, \textit{Attribute}-set and \textit{Value}-set. On the one hand, $\mathcal{A}$ represents the set of elements that characterize a case such as, social and demographic data (e.g., age, gender and current occupation for instance) and, clinical as well as biological data (e.g., existence of hypertension, diabetes, chronic respiratory failure, chronic heart failure, to name a few) \footnote{The complete set of criteria is further detailed in the subsection \ref{subsec:SPop}.}. The set $\mathcal{A}$ is considered common to both $\mathcal{C}_L$ and $\mathcal{C}_U$. On the other hand, ${V}$ (i.e., either one of the sets ${V}_L$ and ${V}_U$) represents a set of vectors related to the considered attributes for every patient:  Let, $v_{a,p}$ refer to the value assigned to the attribute $a\in\mathcal{A}$ for the patient $p \in \mathcal{P}$ (i.e., either one of the sets $\mathcal{P}_L$ and $\mathcal{P}_U$). For the sake of ease of representation, ${V}$ can be seen as a matrix of size\footnote{The notation $|\mathcal{A}|\times|\mathcal{P}|$ represents the value of the product of the cardinal of both sets $\mathcal{A}$ and $\mathcal{P}$.} $|\mathcal{A}|\times|\mathcal{P}|$,  where every cell contains a value $v_{a,p}$. For every attribute $a$, a patient $p$, can either verify the attribute $a$ or not. Consequently, $v_{a,p}$ can only take a binary value in $\{0,1\}$, where $1$ refers to \textit{attribute verified} and $0$ otherwise.  Thus, $V_{p}$ refers to a vector of $|\mathcal{A}|$ binary elements that represents the condition of a patient $p\in\mathcal{P}$ regarding a set of attributes $\mathcal{A}$. 
\par As previously mentioned, the set of patients $\mathcal{P}_L$ considered in $\mathcal{C}_L$ is already labeled. The set of labels $y_{p}$ are stored in a vector ${Y}$.
\vspace{0.2cm}

\par Finally, we can see the decision making process $\pi$ as a function that classifies unlabeled patients in the set $\mathcal{P}_L$ relying on the similarity of the unlabeled patients with the set of labeled patients. Let ${S}$ refer to the vector of labels provided by the decision making engine, where every patient $p\in\mathcal{P}_U$ is assigned a numerical value $s_p\in [0,1]$, such that for every patient $p\in\mathcal{P}_U$:
\begin{align}
s_p=\pi{\left\{\{v_{a,p}\}_{a\in\mathcal{A}},\mathcal{C}_L,{Y}\right\}}
\end{align}
\noindent where $s_p$ quantifies the possible proximity of patient $p$ to the possible classes in $\mathcal{C}_L$. If $s_p$ is a binary value, i.e. $s_p\in \{0,1\}$, the decision making policy $\pi$ is referred to as a hard classification. Otherwise, it is usual to speak of soft classification.   We consider in this paper this latter approach.
\vspace{0.3cm}

\par In the context of CBR, the decision maker assigns a label to new cases depending on their similarity with previously solved cases. The assignment relies on a measure that quantifies the resemblance of the analyzed case with the set of labeled cases. Such decision making approach mimics the decision making process of a physician when dealing with new patients for instance. To do so, the decision maker needs to assess the importance of the different factors as well as the reliability of the cases, i.e. patients, dealt with in the past. 
\par In this paper, the designed CBR relies on a soft $K$-NN algorithm, perhaps one of the most widely used technology in CBR \cite{watson_case-based_1999}. Namely, rather than assigning a label to either classes, we compute a probability of being assigned such labels. Such probability is computed relying on the $K$ most similar patients already labeled. A simple threshold decision making would lead to a hard classification process. 
\par Designing our decision making mechanism requires estimating the distance between patients as well as qualifying the reliability of the labeled patients. These notions are discusses in the next subsections.

\subsection{Similarity Metric and Attributes' weights}
\label{subsec:SMLM}
\par Ideally speaking, similar patients should belong to a same class (registered or not registered). Similar patients usually express similar values to their respective attributes. Equivalently, to the notion of similarity, we can define a distance measure that quantifies the proximity of the new patient to treat with the previously seen patients (i.e. the labeled set of patient). The larger the similarity measure is the smaller becomes the distance.
\par For the sake of simplicity we define, in this paper, the distance measure as follows. Let $p$ and $p'$ denote two patients (label or unlabeled), the distance between these patients is quantified through the measure:
$$
d(p,p')=\sum\limits_{a\in\mathcal{A}}\omega_{a}\left(1-v_{a,p}\oplus v_{a,p'}\right)
$$
\noindent where $\oplus$ refers to the exclusive OR (XOR) operator and such that:
$$\sum\limits_{a\in\mathcal{A}}\omega_{a}=1$$

\par Where, $\omega_{a}$ denotes the weight assigned to attribute $a\in \mathcal{A}$, and the similarity measure appears equal to: $$\sum\limits_{a\in\mathcal{A}}\omega_{a} v_{a,p}\oplus v_{a,p'}$$

\par The weights $\{\omega_a\}_{a\in\mathcal{A}}$ are, usually, not known \textit{a priori}. Therefore, the decision maker needs to acquire that information through a learning process. Thus, relying on the labeled set of cases, the decision maker estimates the impact of the various attributes considered. This step is discussed in Section \ref{sec:LPBLR}, where all required learning steps are detailed.

\subsection{Soft $K$-Nearest Neighbor Algorithm}
\label{subsec:KNN}
$K$-NN Algorithms refer to simple classification techniques that assign labels to new cases depending on their similarity with a reference set of already labeled cases. 
\par Thus, for every new patient $p$ to label, $p\in\mathcal{P}_U$, a $K$-NN algorithm operates through mainly two major steps, the selection step and the fusion step:
\vspace{0.2cm}
\par \textbf{Selection Step:}
\begin{itemize}
\item Computes first the similarity of patient $p$ with patients $p'\in\mathcal{P}_L$.
\item Sort the similar patients $p'\in\mathcal{P}_L$ according to their similarity measure.
\item Select the $K$ most similar patients $p'$.
\end{itemize}
\par \textbf{Fusion Step:} Compute a numerical value that quantifies the proximity of the new case (i.e. Patient $p$) to the set of possible classes in the training set (i.e. $\mathcal{C}_L$).  
\vspace{0.2cm}
\par Depending on this last step, a decision maker can, if needed, assign a label to the new case. Usually a threshold based classifier is used for the assignment process. This latter is however out of the scope of this paper.
\vspace{0.3cm}
\par Let $\mathcal{P}^{*}_K$ refer to the optimal $K$-NN set obtained after the selection step. More specifically $\mathcal{P}^{*}_K$ contains the K labeled patient  -stored in $\mathcal{P}_L$- that have the largest similarity measures with respect to the currently analyzed patient $p\in \mathcal{P}_U$. The fusion step consists in quantifying the possible outcome of the decision making process. 

Finally, the outcome of the decision making process, $s_p$ for a patient $p\in \mathcal{P}_U$ is defined as:
\begin{align}
s_p=\frac{\sum\limits_{p'\in \mathcal{P}^{*}_K}{{\omega_{p'}}{d(p,p')^{-1}}{y_{p'}}}}{\sum\limits_{p'\in \mathcal{P}^{*}_K}{\omega_{p'}}{d(p,p')^{-1}}}
\label{eq:fusion}
\end{align}
\noindent where the set of patients' weights is denoted by the variables $\{\omega_p'\}_{p'\in\mathcal{P}_L}$, and $\{y_p'\}_{p'\in\mathcal{P}_L}$ are the labels assigned to the labeled cases as defined in the subsection \ref{subsec:DMP}. The weights $\{\omega_{p'}\}_{p'\in\mathcal{P}_L}$ are designed to verify: 
$$\sum\limits_{p'\in\mathcal{P}_L}\omega_{p'}=1$$

\vspace{0.1cm}

\par We conclude this subsection discussing, briefly, the settings of the $K$-NN model: i.e., the selection of an appropriate value $K$. Usually, it is not possible to define, \textit{a priori}, the value of the parameter $K$. Thus, a setting phase is necessary to evaluate a satisfactory value with respect to a learning set. 
\par The setting phase consists in three steps. First, a specific subset $\mathcal{C}_S$ of the learning learning set $\mathcal{C}_L$, $\mathcal{C}_S\subset\mathcal{C}_L$, is defined. We refer to this subset as \textit{setting set} in Section \ref{sec:EPR}. Then an evaluation metric that quantifies how well behaves the $K$-NN algorithm on the setting set is computed for the integers ($1, 2, \cdots, K_{max}$) smaller than a specified limit $K_{max}$. Finally the smallest integer $K\in\{1, 2, \cdots, K_{max}\}$ that maximizes the evaluation metric is kept and used on the set $\mathcal{C}_U$ during the learning process. This procedure is further discussed in Section \ref{sec:EPR}.  
\section{Learning Process based on Logistic Regression}
\label{sec:LPBLR}
\par This section deals with the learning phase. As a matter of fact, in order to implement the $K$-NN based CBR, we need to compute, on the one hand, the parameters $\{\omega_a\}_{a\in\mathcal{A}}$ to evaluate the similarity between patients, and on the other hand, the parameters  $\{\omega_p\}_{p\in\mathcal{P}_L}$ in order to evaluate the importance -or contribution- of each patient in $\mathcal{P}_L$. We consider the scenario where the set of parameters is computed once relying on the labeled cases. Then they are exploited to solve new cases.

\subsection{Logistic Regression}
\label{subsec:LR}
\par In a nutshell, Logistic Models (LM) are useful to predict the presence or absence of an outcome or a characteristic based upon the values of a set $\mathcal{A}$ of predictor variables. The methods fits regression model for binary response data relying on the maximum likelihood method \cite{hosmer_applied_2000}. 
More specifically, in this paper we consider the following definition:
\begin{definition}[Logistic Regression]
Let $\mathcal{A}$ denote a set of explanatory variables, $\mathcal{P}_L$ a set of cases, $V$ a binary matrix in $\{0, 1\}^{|\mathcal{A}|\times|\mathcal{P}|}$ such that $\{V\}_{a,p}=v_{a,p}$ with $a\in\mathcal{A}$ and $p\in\mathcal{P}_L$, and finally, let $Y$ refer to a vector of binary expert outcomes  (e.g., \textit{registered} or not \textit{registered}). 
\par LR assumes that there exist an underlying LM that can explain the decision outcomes $Y$ as a logistic function of the matrix $V$ and a vector of regression parameters $\beta \in \mathbb{R}^{|\mathcal{A}|+1}$.
\par Then LR fits the data in $V$ to a logistic function such that for any case $p\in\mathcal{P}$ characterized by a vector of values of the set $\mathcal{A}$:
$$
\widehat{y}_p=\left(1+e^{-(\sum_{a\in\mathcal{A}}v_{a,p}\widehat{\beta}_a+\widehat{\beta}_0)}  \right)^{-1}
$$
\noindent where $\{\{\widehat{\beta}_{a}\}_{\{a\in\mathcal{A}\}}, \widehat{\beta}_0\}$ represent maximum likelihood estimated regression parameters and $\widehat{y}_p$, in $[0,1]$ the estimated prediction outcome for any analyzed case $p$.

\label{def:LR}
\end{definition}
\par In Definition \ref{def:LR}, the regression coefficients reflect the relative influence of predictor factors to define cases' registration on the waiting list. Thus it is natural to take them into account when computing the weights of the attributes $\mathcal{A}$ and the patients $\mathcal{P}_L$ as described in Section \ref{sec:model}. This matter is further detailed in next subsection.

\subsection{Weighting of Attributes and Patients}
\label{subsec:WAP}
\par Significance of each factor, when the regression provides maximum likelihood estimates, is based on the Wald's test defined as follows:

\begin{definition}[Wald Statistic and Weighting of Attributes]
\par Let $\{\widehat{\beta}_{a}\}_{\{a\in\mathcal{A}\}}$ denote a vector of maximum likelihood estimates and $\{\widehat{\sigma}_{a}\}_{\{a\in\mathcal{A}\}}$ their respective maximum likelihood standard deviations. Then Wald's statistic with respect to the attribute $a\in\mathcal{A}$ is defined as:
$$
Wald_a=\frac{\widehat{\beta}_{a}^{2}}{\widehat{\sigma}_{a}^{2}}
$$
\noindent Finally, the vector of weights of attributes, $\{\omega_a\}_{a\in\mathcal{A}}$, is defined such that:
$$
\omega_a=\frac{Wald_a}{\sum_{a'\in\mathcal{A}}Wald_{a'}}
$$
\end{definition} 

\vspace{0.2cm}

\par When dealing with the set of labeled cases $\mathcal{C}_L$, LR introduces a gap between the stored binary outcomes $Y$ and the predicted soft outcomes $\widehat{Y}$. For every $p \in \mathcal{P}_L$, the value of the gap equals $\left(y_p-\widehat{y}_p\right)$. Relying on the definition of Pearson residuals, we introduce the cases' attributes $\{\omega_{p}\}_{\{p\in\mathcal{P}_L\}}$ as follows:
\begin{definition}[Weighting Cases]
Let $p \in \mathcal{P}_L$ denote a labeled case, $y_p$ its label and $\widehat{y}_p$ the logistic regression outcome. Pearson residuals are defined as:
$$
\epsilon_p=\frac{y_p-\widehat{y}_p}{\sqrt{\widehat{y}_p(1-\widehat{y}_p)}}
$$
\noindent where $\epsilon_p$ is assumed to be drawn from a standard normal distribution. Thus $\omega_{p}$ is defined as:
$$
\omega_p=\frac{\mathbb{P}\left(||\epsilon_p||\right)}{\sum_{p'\in\mathcal{P^*}_K}\mathbb{P}\left(||\epsilon_{p'}||\right)}
$$
\noindent where $||\cdot||$ refers to the absolute value function and $\mathbb{P}\left(\cdot\right)$ refers to the probability density function of a standard normal distribution.
\end{definition}
\vspace{0.1cm}
\par We end this section introducing a last notation for the sake of clarity. Usually, many training phases are needed in order to estimated all the parameters of a complete decision making process. In such case, the labeled set $\mathcal{P}_L$ needs to be divided and distributed among the different phases. In this paper, the parameters of both the LM the $K$-NN algorithm need to be learned.  Thus the set $\mathcal{P}_L$ needs to be subdivided into two sets $\mathcal{P}_S$, introduced in previous section, for the sake of the algorithm $K$-NN, and a set $\mathcal{P}_T$, referred to as \textit{training set}, dedicated to the estimations of LM parameters. Finally, $\mathcal{P}_L=\mathcal{P}_T\cup\mathcal{P}_S$ and since $\mathcal{P}_T$ and $\mathcal{P}_S$ must not overlap, i.e., they contain no common cases We can write, to conclude this section, that their intersection is empty: $\mathcal{P}_T\cap\mathcal{P}_S=\emptyset$.
\vspace{0.1cm}
\par The rest of the paper focuses on the implementation, evaluation and interpretation of this methodology.  
\section{Experimental Protocol and Results}
\label{sec:EPR}

\subsection{Data description: Training, Setting and Evaluating sets}
\label{subsec:DD}

\par The initial population included $1647$ patients who began an ESRD treated by dialysis ($652$ ($40\%$) women and $995$ ($60\%$) men). Among them, $350$ i.e., $21\%$, have been registered on the waiting list of renal transplantation in the first year following the start of RRT. 
\par Unfortunately, patients' data with respect to the selected explicative variables (Cf Subsection \ref{subsec:SPop} for further details), were not always complete or fully available. Since, logistic models cannot deal with missing data, we decided to restrict this analysis to a subset of patients with no missing data. 
\par Thus, the study population was reduced to $1137$ patients with complete data, which only represent $70\%$ of the initial population. It is worth mentioning that the general caracteristics of this population remain similar to the original population. As a matter of fact, the population still included a majority of men ($692$ men, $61\%$) and the rate of patients registered on the waiting list remains similar to the original population ($255$ patients, $23\%$).  For the rest of this section, we only focus on the $1137$ patients with complete data. We denote this set of patients $\mathcal{P}$ as introduced in previous sections.

\par Thus, the set of patients $\mathcal{P}$ is such that $|\mathcal{P}|=1137$. For the sake of the experiment,  $\mathcal{P}$ is distributed into two sets:  $\mathcal{P}_L$ and $\mathcal{P}_U$. On the one hand, the set $\mathcal{P}_L$  represents the labeled set that we use for training the LM as well as for setting the parameter $K$ of the $K$-NN algorithm, while on the other hand, we kept a set $\mathcal{P}_U$, considered as the unlabeled data on which we apply our methodology, for the evaluation phase. The labeled set is also partitioned into two sets: $\mathcal{P}_L=\mathcal{P}_T\cup\mathcal{P}_S$. The training set $\mathcal{P}_T$ is dedicated the LM, while the setting set $\mathcal{P}_S$ is used to estimate an appropriate $K$-value of the $K$-NN algorithm. 
\par Finally, the training database, the  setting database and and the evaluation database are built relying on a random sampling for the set population set, such that\footnote{It is worth mentioning that no specific filtering was used to obtain the same number of patients in all three databases. It is a simple coincidence that occurred after discarding patients with incomplete data.}: $$|\mathcal{P}_T|=|\mathcal{P}_S|=|\mathcal{P}_U|=379$$
A Chi-Square test was performed to verify that all three sets share common characteristics. The Chi-Square test showed no significant difference between the three databases. 

\subsection{Experimental Protocol}
\label{subsec:EP}

\begin{figure*}
	\centering
		\includegraphics[width=0.95\textwidth, height=12cm]{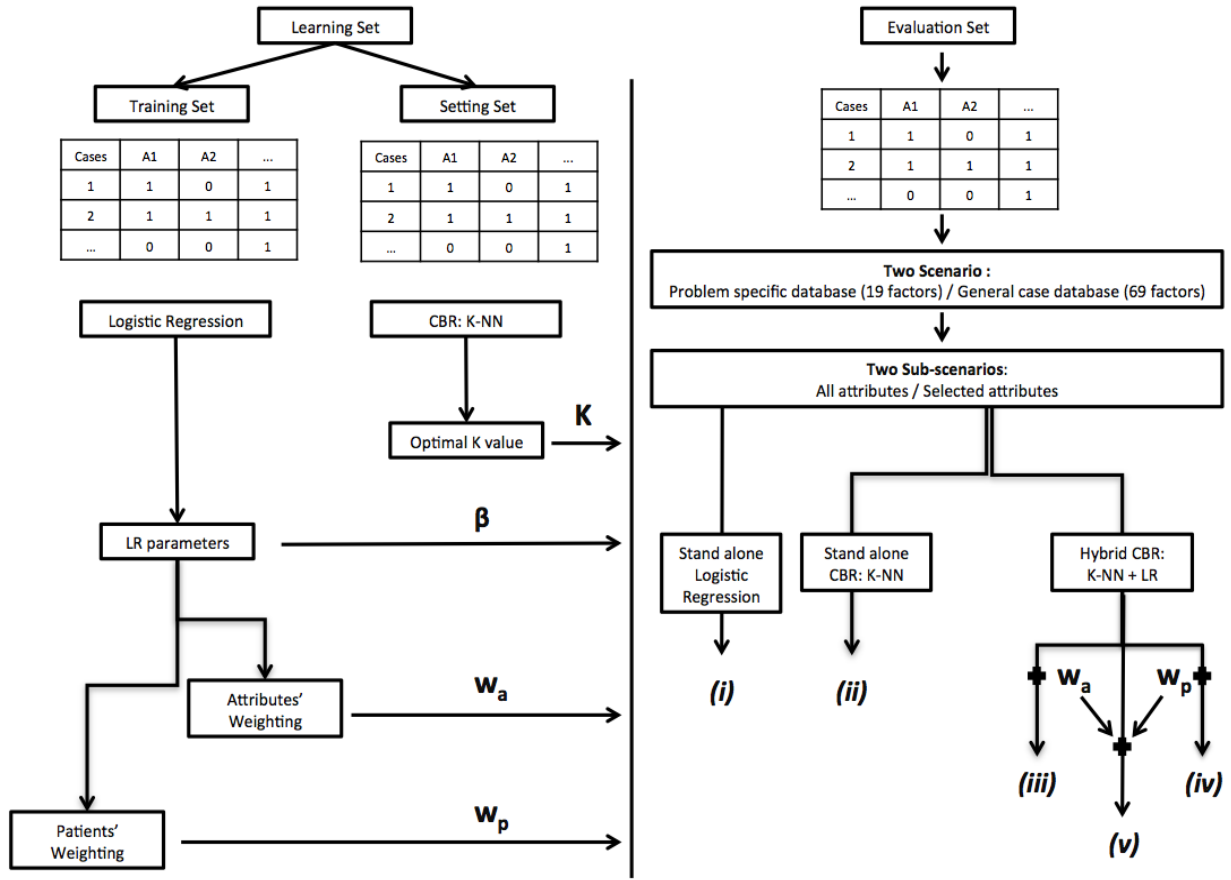}
	\caption{Experimental Protocol. During the learning phase, a training set is used to compute the parameters of a logistic regression model. These parameters enable the computations of the weights of attributes as well as patients' weights. Then a setting set is used to evaluate an optimal $K$ value for the $K$-NN algorithm. Finally all these estimates are exploited to evaluate five decision making algorithms referred to by the indexes \textit{(i)} to \textit{(v)}.}
		\label{fig:Fig1}
\end{figure*}

\par The key aims of this subsection are twofold. On the one hand, we describe the algorithms considered in this experimental section and compare them to the overall approach detailed hereabove. On the other hand, we present the evaluation criteria considered in this paper to assess the quality of the different simulated approaches.

\par As discussed in previous Sections, we consider in this paper the combination of a case based reasoning approach, viz. $K$-NN algorithm, with a logistic regression model. Moreover, in order to enhance its behavior, we suggested several weighing parameters that capture the relevance of the explicative variables and the labeled cases. In order to evaluate the suggested approach, we propose to simulate five different algorithms analyzed within two scenarios.
\par The five algorithms combine different elements described in Sections \ref{sec:model} and \ref{sec:LPBLR}. First we simulate, separately, the two main algorithms describes in previous sections:
\begin{itemize}
\item \textit{(i)} The standalone logistic regression algorithm.  
\item \textit{(ii)} The standalone $K$-NN algorithm (also referred to standalone CBR algorithm in the rest of the paper). 
\end{itemize}
\par Both algorithms were extensively studied and know to be efficient prediction tools. In order to analyze the benefit of weighting the attributes and/or the patients, we start by simulating the standalone versions. Then we progressively add the weighting variables introduced in Subsections \ref{subsec:SMLM} and \ref{subsec:WAP}. This results into three other approaches to consider. Thus, we can enumerate the following algorithms:
\begin{itemize}
\item \textit{(iii)} A $K$-NN with weighted attributes (also referred to as \textit{CBR+$\omega_{a}$} in the simulation results).
\item \textit{(iv)}  A $K$-NN with weighted patients (also referred to as \textit{CBR+$\omega_{p}$} in the simulation results).
\item \textit{(v)}   A $K$-NN with both weighted attributes and weighted patients (also referred to as \textit{CBR+$\omega_{a}$+$\omega_{p}$} in the simulation results). This latter is the suggested approach of this paper. The four other algorithms are used as comparison material.
\end{itemize}

\par All five algorithms are computed within two scenarios: on the one hand, $19$ explicative variables, i.e., attributes, that comply with the general medical model are used. This first scenario analyses the performances of these algorithms when the variables are already reliable form the empirical point of view. On the other hand, $50$ additional attributes randomly defined are considered in the second scenario in order to evaluate the robustness of the simulated algorithms with respect to uncertain models. Namely, the objective is to study the behavior of the prediction tools when the knowledge database contains factors not related to the prediction object.
\par Moreover, in every scenario we evaluate the benefit of automated variable selection for LR before simulating the algorithms. Thus for every scenarios, we describe two sub-scenarios. We refer to them in simulations as the sub-scenarios \emph{Prediction using all attributes} and \emph{Prediction using selected attributes}\footnote{The selection procedure can be referred to as \textit{stepwise selection}.}. All scenarios and algorithms are summarized in Figure \ref{fig:synthese}.

\par All performance results are presented in terms of the receiver operating characteristic curve (AUC). In order to compute confidence intervals of AUC results, a bootstrap resampling procedure is performed \cite{skalska_web-bootstrap_2006}. Thus, the probability distribution of AUC statistic is simulated by $500$ random samples from the original evaluation database. Then a specific non parametric Monte Carlo AUC estimator, $\overline{AUC}$, is computed. The chosen estimator is a non biased AUC estimator such that: 
$$
\overline{AUC}=\frac{\sum_{b=1}^{k} AUC_{b}}{k}
$$
\noindent where the index $b$ refers to the bootstrap iteration and $k$ is the total number of iterations ($k=500$ in this case).
\par We computed the performance evaluation estimates such that the confidence intervals limits are the $2.5$ and $97.5$ percentiles of the  $\overline{AUC}$ distribution. 
\vspace{0.1cm}
\subsection{Computational Tools}
\label{subsec:CT}
\par All computations involved in this study, including LM based regression and CBR algorithms, were performed on the free software environment `R' \footnote{ Version 2.12.2 GUI 1.36 Leopard build 32-bit for Mac OS X \cite{r_development_core_team_r:_2011}.}. 
\par More specifically, we relied on the package `stats' (version 2.12.2) to implement logistic regression. As a matter of fact, it allows modeling generalized linear models thanks to the `glm' function. Then, the functions `Anova' and `summary' enabled the estimation of our LM parameters. Finally, the function `step' was used for selecting LR variables relying on a stepwise procedure and on Akaike's criterion. 
\par Related to CBR algorithms, we designed our specific functions using the programming language of the R user interface to ensure calculation of similarity measures, selection of nearest neighbors, prediction of probability to be registered, and classification of cases. 

\subsection{Results}
\label{subsec:results}

\begin{table*}[t]
\centering
\includegraphics[width=1\textwidth]{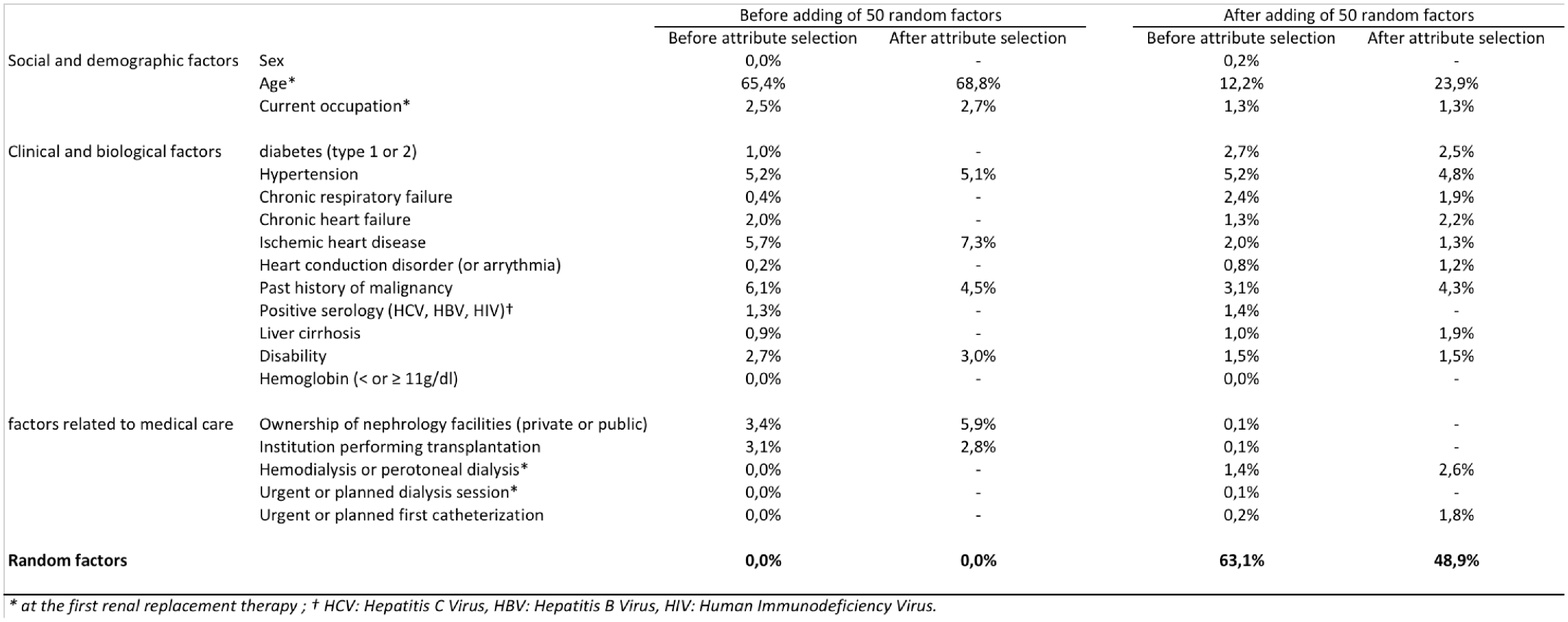}
\caption{List of attributes and weights used by $K$-Nearest Neighbors algorithms before and after adding the 50 random attributes, and before and after the stepwise selection procedure of attributes.}
\label{tab:Tab1}
\end{table*}

\par Table \ref{tab:Tab1} shows the weights of attributes calculated from the Wald statistics using the regression coefficient estimations of the LM, as defined in Subsection \ref{subsec:WAP}, and their respective standard deviations. Both sub-scenarios, summarized in Figure \ref{fig:Fig1}, are considered where estimations are conducted after (or without) a stepwise selection procedure on the set of explicative variables (viz, attributes). The results of Table \ref{tab:weights} consider first the case database with only $19$ attributes relevant to our problem (referred to as \textit{before adding of $50$ random factors}). Then, $50$ random attributes are added and the computations of both sub-scenarios are once again repeated. 
\par As expected, the attributes have a different impact on the registration. Their respective impact reflects on the performance of the $K$-NN algorithm through the values of the weights of attributes. 
\par When only the $19$ relevant factors are considered and without a stepwise selection procedure, the most relevant predictive factors seem to be: age, hypertension, ischemic heart disease, past history of malignancy, ownership of nephrology facilities and follow-up in institution performing renal transplantation. It is worth noting that age and past history of malignancy are the only factors with a significant Wald test value. After the stepwise selection procedure, LM kept the same eight predictive factors where age, hypertension, ischemic heart disease and ownership of nephrology facilities showed a significant Wald test value. 
\par We can notice that the logistic regression performed in this study showed results equivalent to those described in recent literature \cite{bayat_medical_2006,ravanan_variation_2010}.  We used both medical and non-medical predictive factors of transplant registration. As mentioned in Subsection \ref{subsec:DomApp}, non-medical factors might not be relevant for clinical practice ; however our main objective is to discuss the efficiency of a new computational $K$-NN and not to meet concrete decision-making applications. 
\par \textit{Age} in this kind of application field is, with no surprise, one of the most relevant clinical factors. As it could be expected, it showed a very high weight level compared to other factors. This fact might limit the results of the study. Nevertheless, since we need to design a decision-making process that performs automatically, we decided to keep the factor \textit{age} within the discriminating factors in LM and $K$-NN algorithms.

\vspace{0.1cm}

\begin{figure*}[htbp]
\centering
	\subfloat[Prediction using all attributes]{\includegraphics[scale=0.3]{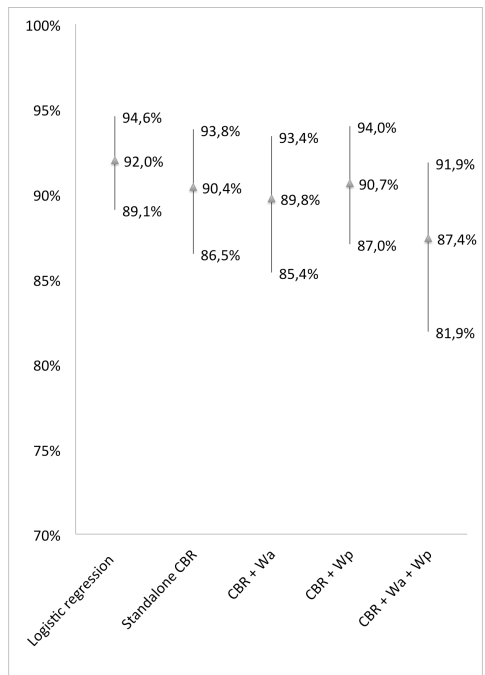}\label{fig:Fig2}}\qquad
	\subfloat[Prediction using selected attributes]{\includegraphics[scale=0.3]{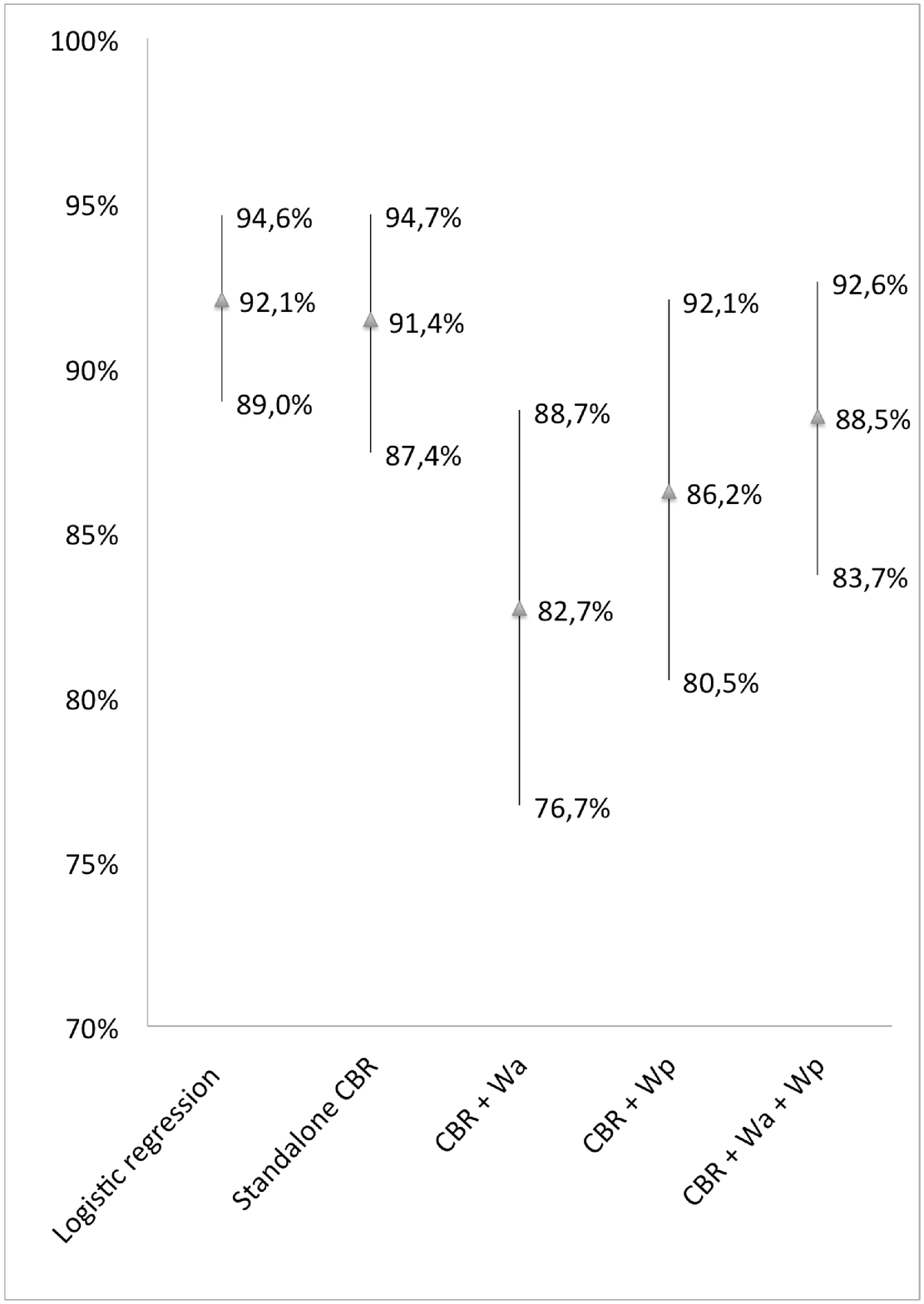}\label{fig:Fig3}}
\caption{Prediction results before adding $50$ random attributes.}
\label{fig:PredPart1}
\end{figure*}

\begin{figure*}[htbp]
\centering
	\subfloat[Prediction using all attributes]{\includegraphics[scale=0.3]{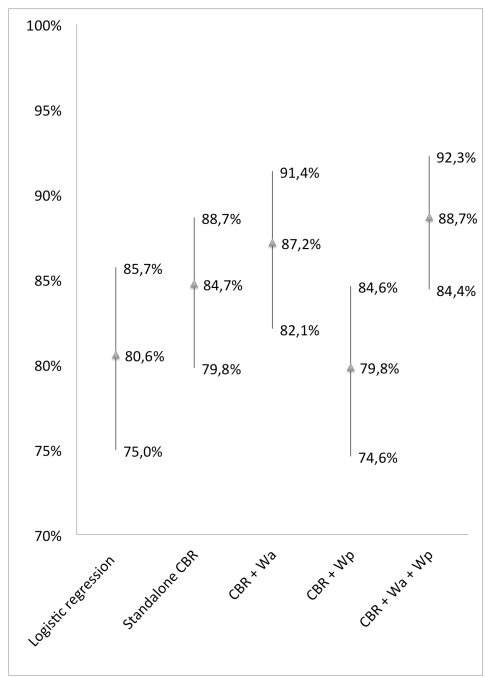}\label{fig:Fig4}}\qquad
	\subfloat[Prediction using selected attributes]{\includegraphics[scale=0.3]{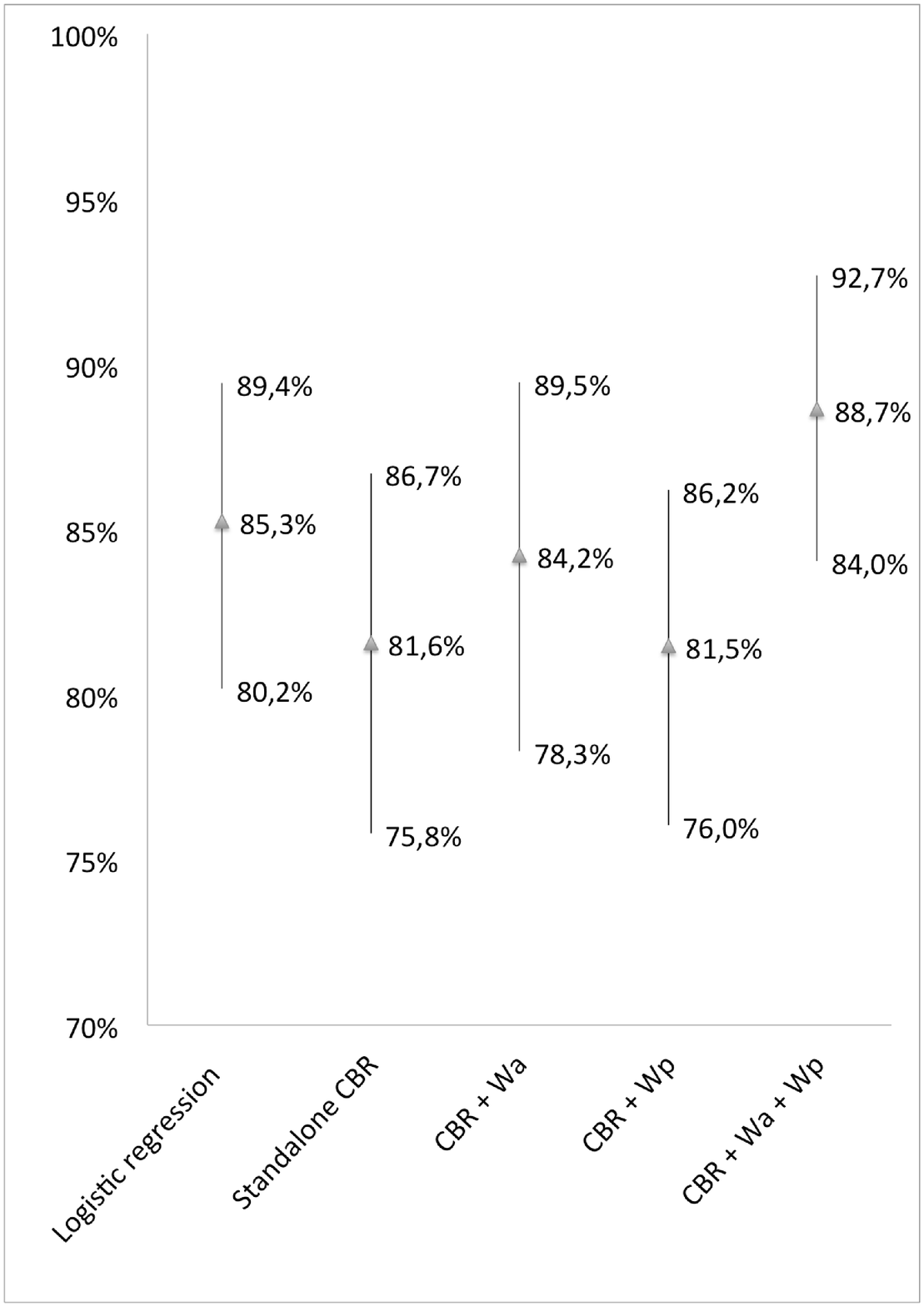}\label{fig:Fig5}}
\caption{Prediction results after adding $50$ random attributes.}
\label{fig:PredPart2}
\end{figure*}

\par After adding $50$ random factors, estimations from the LM and the weights of attributes showed a significant change. As a matter of fact, the weight of \textit{age} at the first RRT, for example, decreased from $65\%$ and $69\%$, respectively before and after stepwise attribute selection, to $12\%$ and $24\%$ in the protocol arm including the random factors. Overall, the role of both the socio-demographic factors and the factors related to medical care decreased after the introduction of random factors, while the role of clinical and biological factors remained stable. The decrease of the values of sociodemographic factors' weights and factors related to medical care happened in favor of random factors that kept a significant weight on prediction despite the selection of the attributes by a stepwise selection procedure. As expected, adding random factors creates an artifact in the definition of the relevant factors and the course of the prediction procedure. This artefact help us assess the robustness of LM combined with K-NN algorithms which is discussed in the rest of this Section.

\vspace{0.1cm}

\par Figures \ref{fig:PredPart1} and \ref{fig:PredPart2} show prediction results performed by the LM and the CBR methods using the $K$-NN standalone, the $K$-NN with weighting of either attributes or patients, and using the $K$-NN with weighting of both patients and attributes; respectively before and after adding $50$ random attributes (as summarized in Figure \ref{fig:Fig1}). 

\par First of all, we evaluate the performance of the algorithms in the ideal case with no artifact, i.e., only the $19$ relevant attributes are considered. In this context, results show that predictions provided by LM and standalone CBR methods tend to be more powerful than methods combining $K$-NN and LM. This is not a surprise as both LM and $K$-NN are known to be quite efficient when the attributes are relevant.
\par Right sub-figure in Figure \ref{fig:PredPart1} shows the performances of the tested algorithm in the ideal case with no artifact, however a pre-selection of the attributes in conducted before computing the algorithms. We notice that their performances do not significantly change except for the algorithm referred to as \textit{CBR+$\omega_a$} (viz, $K$-NN with weighted attributes). As a matter of fact, we notice that this latter suffers a significant performance decrease. Since a stepwise selection of the attributes is conducted before launching the algorithm, i.e., before weighting the attributes and computing the $K$-NN algorithm, we can conclude that the stepwise attribute selection might discard some of the attributes that seem to have a significant impact when the attributes are weighted later. 
\par Then, a similar evaluation is performed after adding $50$ random attributes that, usually, are not considered as relevant. In such a scenario, the standalone LM and $K$-NN could suffer difficulties as the context is not optimally chosen to tune their performances. This is indeed observed in Figure \ref{fig:PredPart2} where the performances of standalone LM and $K$-NN degrade significantly.
\par One of the most interesting results through out Figures \ref{fig:PredPart1} and \ref{fig:PredPart2} is the robustness of the combination of LM and CBR when both attributes and patients are weighted. As a matter of fact, in all scenarios, with or without artifact, with or without stepwise attribute selection, the algorithm referred to as \textit{CBR+}$\omega_a \textit{+}\omega_p$ performs in a consistent way. It provides for all scenarios a prediction rate around $88\%$ ; whereas all other algorithms, tested in this paper, seem to suffer at one point or another. This robustness offers a performance guaranty. This latter might prove to be less efficient than others in some specific scenarios, however since in realistic scenarios it is usually impossible to tell \textit{a priori} wheather there is an artifact or not, choosing the algorithm that combines both weighted attributes and weigthed cases seems to be a cautious choice.

\section{Related Works and Perspectives}
\label{sec:Related}
\par Logistic regression analyses are widely used in medical research, however it is more commonly reserved for determining prognostic factors than for predicting disease. To our knowledge, no study evaluates prediction of access to the french renal transplant waiting list by LM. 
\par Bayat \textit{et al} invested the issue in two recent publications using a Bayesian network and a decision tree method \cite{bayat_comparison_2009}. They do not present any AUCs, thus it is not possible to directly compare their results with ours. However, they conclude both methods have very high predictive performances and age is the most important factor for predicting access to the waiting list, which is coherent with our results.
\par Similarly, Chuang compared several classifiers including LM and CBR methods to predict presence of liver disease \cite{chuang_case-based_2011}. For the author, results related to CBR methods testify to the solid diagnosis capacity of CBR in examining healthy data. Our results support this conclusion since we have shown that CBR method present predictive performances equivalent to those obtained by LM. This paper shows however that it is true only if the considered attributes are well chosen and reliable regarding the problem to solve.

\vspace{0.2cm}

\par Nugent \textit{et al} presented the first association between CBR and LM in 2009 with a methodology called KLEF for \emph{Knowledge - Light Explanation Framework} \cite{nugent_gaining_2009}. The method describes how gaining high-level knowledge by a \emph{top-down} mechanism using logistic regression. LM is used a posteriori to define one nearest neighbor from cases retrieved by a $K$-NN algorithm. 
\par LM in the present study was used differently. As a matter of fact, the logistic model was directly fitted from the overall knowledge database. Information from LM was a direct contribution to compute similarity measures and classification probabilities. This latter approach is described by Stahl \textit{et al} as a \emph{bottom-up} mechanism \cite{stahl_defining_2002}. 

\vspace{0.2cm}

\par To the best of our knowledge, only two publications describe methods similar to our hybrid approach. The first one is applied to breast cancer diagnosis (Huang \textit{et al} \cite{huang_usage_2010}) and the second one is applied to the diagnosis of liver disease (Chuang \cite{chuang_case-based_2011}). 
\par In Chuang\rq{}s paper, CBR methodology is different from the one applied in the present study. As a matter of fact, similarity measures are performed separately for cases with and without liver disease. Thus, in Huang\rq{}s paper, similarity computation is performed through a $K$-NN algorithm as in the present work. However, LM is only used for defining the most relevant factors and to compute attribute weights. 
\par In the present study, LM is also used to perform attribute selection and attribute weighting. However, we proposed in addition to introduce Pearson residuals to weight the cases in the design of our $K$-NN algorithm. In our opinion, Pearson residuals based case weighting participate, with attribute weighting, to the cases' description ans specification when defining problem-specific knowledge \cite{bergmann_utility-oriented_2001}.Thus, LM defines an archetype of registered and not registered patients in the knowledge database, and LM residuals reflect the adequacy of each patients with regard to the archetype. Relying only on logistic regression coefficients or stepwise selection to define the cases as well as the problem utility would consider that all patients match perfectly the LM archetype. We know for a fact that it is not true. Hence, computing specific weights for each case, relying on LM residuals, appears as an attempt to correct of that approximation. To the best of our knowledge, this is the first time that such an approach is discussed in the literature. 


\par As for Chuang\rq{}s paper, the author points out classification improvements relying on Hybrid CBR approach compared to a standalone CBR. Huang's publication also compares several kinds of hybrid approaches: a neural network with or without fuzzy logic and two hybrid CBR systems, one combining CBR with a decision tree and one combining CBR with LM. The neural networks show superior performances, but the authors emphasized rapidity of cases retrieval and the more easily interpretable results of CBR methodology. 
\par In the present study, the CBR hybrid approaches did not show significant improvements for patient classification, compared to standalone CBR approach. However, the hybrid CBR system combing both attribute weighting and case weighting seems to be very robust to artifacts in the database that might occur in all realistic scenarios. From our point of view, this interesting observation provides new perspectives for future CBR system, particularly for integrating CBR systems into large and unspecific knowledge database as electronic health records \cite{bichindaritz_case-based_2006-1,van_den_branden_integrating_2011}.

\vspace{0.2cm}

Finally, we join Huang \textit{et al}'s opinion as we believe that CBR is an explicit problem solving methodology. We believe that an association between LM and CBR systems improves comprehensiveness of problem-solving processing. This latter provids the users with more reliable information about relevant decision factors and case utility. Thus, the integration of bio-satistical analyses, widely used in the medical research, may also participate in the dissemination and development of CBR decision support for medical practice.

\section{Conclusion}
\label{sec:conc}

\par In the paper, we presented and detailed different ways of coupling $K$-NN algorithm and LR. We have used logistic modeling in order to perform selection and weighting of cases' features, and a new methodology have been proposed to define cases' utility using residuals of the LR. LR herein worked as an automated bottom-up procedure to define problem-specific similarity measures, and we have showed that it could improve algorithms of case retrieval and optimize reuse of cases, and at the same time it could improve CBR performance ans robustness, especially when facing unspecific knowledge case databases.
\par In our opinion, CBR integration in medical decision support is not only dependent of our ability to introduce practical- and patient-oriented data elements in problem-solving procedure, even though they are essential for decision making in medical practice, but also on their ability to be fully integrated into medical reasoning processes. The complete hybrid approach we suggested and discussed, could thus also participate to meet both requirements.

\section*{Acknowledgment}
\noindent The authors would like to thank the \emph{French registry REIN} and the \emph{Agence de la Biom\'edecine} for the contribution of data.

\section*{Conflict of Interest Statement} \vspace{0.2cm}
\noindent The authors declares that there is no conflict of interest.

\bibliographystyle{vancouver}
\bibliography{refs}

\begin{thebibliography}{10}

\bibitem{bichindaritz_advances_2011}
Bichindaritz I, Montani S.
\newblock Advances in case-based reasoning in the health sciences.
\newblock Artificial Intelligence in Medicine. 2011 Feb;51(2):75--79.
\newblock {PMID:} 21397782.
\newblock Available from:
  \url{http://www.sciencedirect.com/science/article/pii/S0933365711000029}.

\bibitem{pantazi_case-based_2004}
Pantazi SV, Arocha JF, Moehr JR.
\newblock Case-based medical informatics.
\newblock {BMC} Medical Informatics and Decision Making. 2004 Nov;4:19.
\newblock {PMID:} 15533257 {PMCID:} {PMC544898}.
\newblock Available from:
  \url{http://www.ncbi.nlm.nih.gov/pmc/articles/PMC544898/}.

\bibitem{aamodt_case-based_1994}
Aamodt A, Plaza E.
\newblock Case-based reasoning: Foundational issues, methodological variations,
  and system approaches.
\newblock {AI} communications : the European journal on artificial
  intelligence. 1994 Mar;7(1):39--59.

\bibitem{bichindaritz_case-based_2006-1}
Bichindaritz I, Marling C.
\newblock Case-based reasoning in the health sciences: What's next?
\newblock Artificial Intelligence in Medicine. 2006 Feb;36(2):127--135.
\newblock {PMID:} 16459064.
\newblock Available from:
  \url{http://www.sciencedirect.com/science/article/pii/S0933365705001247}.

\bibitem{bergmann_experience_2002}
Bergmann R.
\newblock Experience Management : Foundations, Development Methodology, and
  Internet-Based Applications.
\newblock {(Lecture} Notes in Computer Science, {ISSN} 0302-9743 ; 2432).
  Berlin, Heidelberg : Springer-Verlag Berlin Heidelberg: Springer e-books;
  2002.
\newblock Available from:
  \url{http://www.springerlink.com/content/jdcgyq5wd4eh/?MUD=MP}.

\bibitem{bergmann_utility-oriented_2001}
Bergmann R, Richter MM, Schmitt S, Stahl A, Vollrath I.
\newblock Utility-oriented matching: A new research direction for case-based
  reasoning.
\newblock In: Proceedings of the 1st Conference on Professional Knowledge
  Management; 2001. p. 264--274.

\bibitem{couchoud_renal_2006}
Couchoud C, Stengel B, Landais P, Aldigier JC, De~Cornelissen F, Dabot C,
  et~al.
\newblock The renal epidemiology and information network {(REIN):} a new
  registry for end-stage renal disease in France.
\newblock Nephrology Dialysis Transplantation. 2006 Feb;21(2):411--418.
\newblock Available from: \url{http://ndt.oxfordjournals.org/content/21/2/411}.

\bibitem{bayat_medical_2006}
Bayat S, Frimat L, Thilly N, Loos C, Brian{\c c}on S, Kessler M.
\newblock Medical and non-medical determinants of access to renal transplant
  waiting list in a French community-based network of care.
\newblock Nephrology Dialysis Transplantation. 2006 Oct;21(10):2900--2907.
\newblock Available from:
  \url{http://ndt.oxfordjournals.org/content/21/10/2900}.

\bibitem{bayat_modelling_2008}
Bayat S, Cuggia M, Kessler M, Brian{\c c}on S, Le~Beux P, Frimat L.
\newblock Modelling access to renal transplantation waiting list in a French
  healthcare network using a Bayesian method.
\newblock Studies in health technology and informatics. 2008;136:605--610.

\bibitem{bayat_comparison_2009}
Bayat S, Cuggia M, Rossille D, Kessler M, Frimat L.
\newblock Comparison of Bayesian Network and Decision Tree Methods for
  Predicting Access to the Renal Transplant Waiting List.
\newblock Studies in health technology and informatics. 2009;150:600--604.

\bibitem{fritsche_practice_2000}
Fritsche L, Vanrenterghem Y, Nordal KP, Grinyo JM, Moreso F, Budde K, et~al.
\newblock Practice variations in the evaluation of adult candidates for
  cadaveric kidney transplantation: a survey of the European Transplant
  Centers.
\newblock Transplantation. 2000 Nov;70(10):1492--1497.

\bibitem{oniscu_equity_2003}
Oniscu GC, Schalkwijk AA, Johnson RJ, Brown H, Forsythe JL.
\newblock Equity of access to renal transplant waiting list and renal
  transplantation in Scotland: cohort study.
\newblock {BMJ} : British Medical Journal. 2003 Nov;327(7426):1261.
\newblock {PMID:} 14644969 {PMCID:} {PMC286245}.
\newblock Available from:
  \url{http://www.ncbi.nlm.nih.gov/pmc/articles/PMC286245/}.

\bibitem{jeffrey_comparison_2005}
Jeffrey RF, Akbani H, Scally AJ, Peel R.
\newblock Comparison of transplant listing strategy in two renal dialysis
  centers within a regional transplant alliance.
\newblock Clinical nephrology. 2005 Dec;64(6):438--443.
\newblock {PMID:} 16370156.

\bibitem{ravanan_variation_2010}
Ravanan R, Udayaraj U, Ansell D, Collett D, Johnson R, {O'Neill} J, et~al.
\newblock Variation between centres in access to renal transplantation in {UK:}
  longitudinal cohort study.
\newblock {BMJ} {(Clinical} Research Ed). 2010 Jul;341:c3451--c3451.
\newblock {PMID:} 20647283.
\newblock Available from: \url{http://www.ncbi.nlm.nih.gov/pubmed/20647283}.

\bibitem{watson_case-based_1999}
Watson I.
\newblock Case-based reasoning is a methodology not a technology.
\newblock Knowledge-based systems. 1999 Oct;12(5-6):303--308.

\bibitem{hosmer_applied_2000}
Hosmer DW, Lemeshow S.
\newblock Applied logistic regression.
\newblock 2nd ed. Wiley series in probability and statistics. Texts and
  references section. New York ; Chichester ; Weinheim [etc.]: John Wiley \&
  Sons, Inc. A Wiley-Interscience Publication; 2000.

\bibitem{skalska_web-bootstrap_2006}
Skalsk{\'a} H, Freylich V.
\newblock Web-bootstrap estimate of area under {ROC} curve.
\newblock Austrian journal of statistics. 2006;35(2\&3):325--330.

\bibitem{r_development_core_team_r:_2011}
Team RDC. R: A Language and Environment for Statistical Computing, R Foundation
  for Statistical Computing, Vienna, Austria.. {ISBN} 3-900051-07-0; 2011.
\newblock Available from: \url{http://www.R-project.org/}.

\bibitem{chuang_case-based_2011}
Chuang CL.
\newblock Case-based reasoning support for liver disease diagnosis.
\newblock Artificial Intelligence in Medicine. 2011 Sep;53(1):15--23.
\newblock Available from:
  \url{http://www.sciencedirect.com/science/article/pii/S0933365711000728}.

\bibitem{nugent_gaining_2009}
Nugent C, Doyle D, Cunningham P.
\newblock Gaining insight through case-based explanation.
\newblock Journal of Intelligent Information Systems. 2009 Jun;32(3):267--295.
\newblock Available from:
  \url{http://www.springerlink.com/content/p2412nm508962t10/}.

\bibitem{stahl_defining_2002}
Stahl A.
\newblock Defining Similarity Measures: Top-Down vs. Bottom-Up.
\newblock In: Craw S, Preece A, editors. Advances in Case-Based Reasoning. vol.
  2416 of Lecture Notes in Computer Science. Springer Berlin / Heidelberg;
  2002. p. 91--119.
\newblock Available from:
  \url{http://www.springerlink.com/content/l6d1x82e67mt4fkb/abstract/}.

\bibitem{huang_usage_2010}
Huang ML, Hung YH, Lee WM, Li RK, Wang TH.
\newblock Usage of Case-Based Reasoning, Neural Network and Adaptive
  Neuro-Fuzzy Inference System Classification Techniques in Breast Cancer
  Dataset Classification Diagnosis.
\newblock Journal of Medical Systems. 2010 May;{PMID:} 20703710.
\newblock Available from: \url{http://www.ncbi.nlm.nih.gov/pubmed/20703710}.

\bibitem{van_den_branden_integrating_2011}
Van~den Branden M, Wiratunga N, Burton D, Craw S.
\newblock Integrating case-based reasoning with an electronic patient record
  system.
\newblock Artificial Intelligence in Medicine. 2011 Feb;51(2):117--123.
\newblock Available from:
  \url{http://www.sciencedirect.com/science/article/pii/S0933365710001429}.

\end{thebibliography}


\end{document}